\documentclass[conference]{IEEEtran}

\usepackage[numbers]{natbib}
\usepackage{multicol}
\usepackage[bookmarks=true,colorlinks=true]{hyperref}
\usepackage{times}
\usepackage{epsfig}
\usepackage{graphicx}
\usepackage{amsmath}
\usepackage{booktabs}
\usepackage{siunitx}
\usepackage[utf8x]{inputenc}
\usepackage{multirow}
\usepackage{amssymb}
\usepackage{lib}
\usepackage{cite}
\usepackage{enumitem}
\usepackage[dvipsnames]{xcolor}
\usepackage[hang,flushmargin]{footmisc}
\usepackage[labelfont=bf,size=small]{caption}
\newcommand{\linebreakand}{
  \end{@IEEEauthorhalign}
  \hfill\mbox{}\par
  \mbox{}\hfill\begin{@IEEEauthorhalign}
}
\newcommand{\insertW}[2]{\IfFileExists{#2}{\includegraphics[width=#1\textwidth]{#2}}{\includegraphics[width=#1\textwidth]{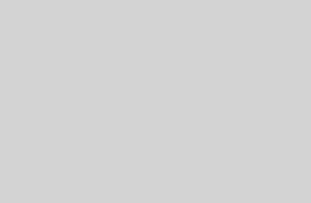}}}
\newcommand{\insertWL}[2]{\IfFileExists{#2}{\includegraphics[width=#1\linewidth]{#2}}{\includegraphics[width=#1\linewidth]{img/blank.png}}}
\newcommand{\insertH}[2]{\IfFileExists{#2}{\includegraphics[height=#1\textwidth]{#2}}{\includegraphics[height=#1\textwidth]{img/blank.png}}}
\newcommand{\insertHW}[3]{\IfFileExists{#3}{\includegraphics[height=#1\textwidth,width=#2\textwidth]{#3}}{\includegraphics[height=#1\textwidth,width=#2\textwidth]{img/blank.png}}}
\newcommand{\insertHWL}[3]{\IfFileExists{#3}{\includegraphics[height=#1\textwidth,width=#2\linewidth]{#3}}{\includegraphics[height=#1\textwidth,width=#2\linewidth]{img/blank.png}}}

\newcommand{\rgb}{RGB\xspace}
\newcommand{\lidar}{LiDAR\xspace}
\newcommand{\CF}{CropFollow\xspace}

\begin{document}

\title{Learned Visual Navigation for Under-Canopy Agricultural Robots}

\author{\authorblockN{Arun Narenthiran Sivakumar$^1$ \quad
Sahil Modi$^2$ \quad
Mateus Valverde Gasparino$^1$ \quad
Che Ellis$^3$ \quad
\\Andres Eduardo Baquero Velasquez$^1$ \quad
Girish Chowdhary$^{*,1,2}$ \quad
Saurabh Gupta$^{*,4}$}

\authorblockA{$^1$Department of Agricultural and Biological Engineering, University of Illinois at Urbana-Champaign (UIUC)} \authorblockA{$^2$Department of Computer Science, UIUC, $^4$Department of Electrical and Computer Engineering, UIUC,}
\authorblockA{$^3$EarthSense Inc.}
}

\maketitle
\begin{abstract}
   This paper describes a system for visually guided autonomous navigation of under-canopy farm robots. Low-cost under-canopy robots can drive between crop rows under the plant canopy and accomplish tasks that are infeasible for over-the-canopy drones or larger agricultural equipment. However, autonomously navigating them under the canopy presents a number of challenges: unreliable GPS and \lidar, high cost of sensing, challenging farm terrain, clutter due to leaves and weeds, and large variability in appearance over the season and across crop types. We address these challenges by building a modular system that leverages machine learning for robust and generalizable perception from monocular \rgb images from low-cost cameras, and model predictive control for accurate control in challenging terrain. Our system, CropFollow, is able to autonomously drive 485 meters per intervention on average, outperforming a state-of-the-art \lidar based system (286 meters per intervention) in extensive field testing spanning over 25~km.
\end{abstract}

\IEEEpeerreviewmaketitle

\section{Introduction}
\let\thefootnote\relax\footnote{
\noindent Project website with data and videos: \url{https://ansivakumar.github.io/learned-visual-navigation/}. \\
Correspondence to \texttt{\{av7,girishc\}@illinois.edu}. \\
$^*$Girish Chowdhary and Saurabh Gupta contributed equally and are listed alphabetically.}
This paper describes the design of a visually-guided navigation system for compact, low-cost, under-canopy agricultural robots for commodity row-crops (corn, soybean, sugarcane etc), such as that shown in \figref{overview}. Our system, called \CF, uses monocular RGB images from an on-board front-facing camera to steer the robot to autonomously traverse in between crop rows in harsh, visually cluttered, uneven, and variable real-world agricultural fields. Robust and reliable autonomous navigation of such under-canopy robots has the potential to enable a number of practical and scientific applications: High-throughput plant phenotyping~\citep{mueller2017robotanist,kayacan2018embedded,young2019design,xu2018development, stager2019design,gao2018novel}, ultra-precise pesticide treatments, mechanical weeding~\citep{McAllister2020RSS}, plant manipulation~\citep{Chowdhary2019MDPI,Uppalapati2020RSS}, and cover crop planting~\citep{vougioukas2019agricultural, velasquez2020reactive}. Such applications are not possible with over-canopy larger tractors and UAVs, and are crucial for increasing agricultural sustainability~\citep{shamshiri2018research,foley2011solutions}.

Autonomous row-following is a foundational capability for robots  that need to navigate between crop rows in agricultural fields. Such robots cannot rely on RTK (Real-Time Kinematic)-GPS \citep{farrell2008aided} based methods which are used for over-the-canopy autonomy (e.g. for drones, tractors, and combine harvesters) because of GPS signal attenuation and multi-path errors. The under-canopy row-following task consists of detecting and following the rows of crop, by determining the distance from the rows and the angle relative to the row, and using this to track specified row-relative pose. In a typical 80 acre land-parcel in row-crops, the rows are about 400 meter long and full of visual clutter. The crop rapidly grows during the growing season, rendering a constantly changing visual environment. Therefore, autonomous navigation of under-canopy robots has remained a challenging and open problem. \lidar is known to work under the canopy and can return geometric information~\citep{higuti2019under}. However, \lidar is costly, and it does not capture semantic information. For example, \lidar cannot directly distinguish whether observed occupancy corresponds to untraversable obstacles (actual crop plant stalk), or traversable obstacles (hanging leaves, weeds, uneven terrain). This fundamentally limits \lidar based methods from estimating distance and angle from the row, leading to low robustness of autonomy, as reported by low distance-between-interventions~\citep{higuti2019under}. This motivates our use of richer sensing and lower-cost modalities in the form of RGB images.

\begin{figure}
\includegraphics[width=1\linewidth]{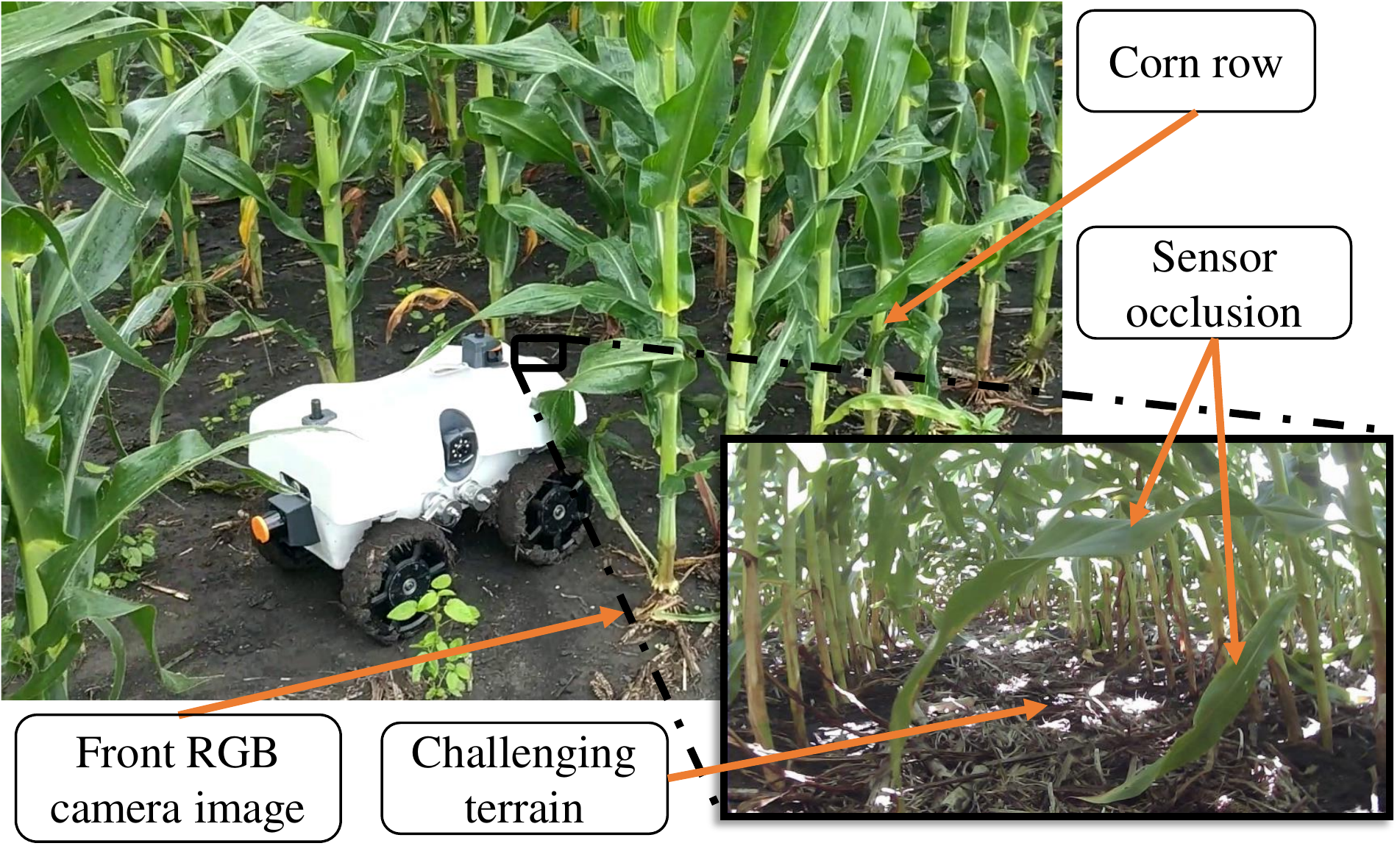}
\caption{\CF is an autonomous navigation system for under-canopy agriculture robots. It uses \rgb images from a front-facing camera to output steering commands to drive the robot in crop rows.}
\figlabel{overview}
\end{figure}

Using \rgb images for under-canopy navigation however has proven to be non-trivial and has become a primary bottleneck for under-canopy robotics. Importance of semantics precludes the use of traditional methods that infer geometry from monocular RGB image streams \citep{mur2015orb,furukawa2009accurate}. Visual variability during the day and the season limits heuristic based crop-lane detection algorithms, and visual similarity results in positional drift with SLAM algorithms\citep{xue2012variable}. It is clear therefore, that the high variability and clutter in the agricultural environment necessitates the use of learning. However, the lack of large-scale datasets, the difficulty of collecting field data, lack of a clear reward signal, and the infeasibility of building a simulator for this task, makes it challenging to employ machine learning.

Our contribution in this paper is a field-validated modular vision based crop-row following system to overcome the above challenges. We term this system CropFollow, as it provides the foundational row-following capability to small, low-cost robots. Our system  decouples  perception and control. The perception system uses monocular \rgb image from the on-board camera to estimate row-relative robot pose. It does so by directly estimating the robot's relative heading to the row (measured as the angle the robot makes with the row direction), and robot's placement in row (measured as the ratio of distance from the left row to inter-row separation). These data are fused with inertial measurements using a Bayesian sensor fusion system (Extended Kalman filter (EKF)), and utilized to generate row-following control in terms of desired angle and speed for staying in the center of the row using a nonlinear robust controller (Model Predictive Control (MPC)).  The ability to directly predict relative heading and distance from monocular RGB images is one key novelty of our approach, and has key efficiency and robustness benefits: the approach avoids having to first detect the plants (which can be many) \citep{gu2020path}, or explicitly segmenting the ground from plants (which is highly challenging with more clutter in the environment) \citep{xue2012variable}. Our presented system is able to successfully traverse crop rows regardless of the crop's growth stage. In field trials of about 25 kilometers, our system required fewer interventions than a \lidar based system \citep{Velasquez2021Multisensorfusion}(485 meters per intervention \vs 286 m), while at the same time cutting down sensing cost by 50$\times$. In offline experiments, we find that the proposed perception models generalize well to new crops. These results clearly establish that our modular visual navigation system enables vision based autonomy for under-canopy field robots.

\section{Related Work}
\seclabel{related}

\begin{figure*}

\includegraphics[width=1\linewidth]{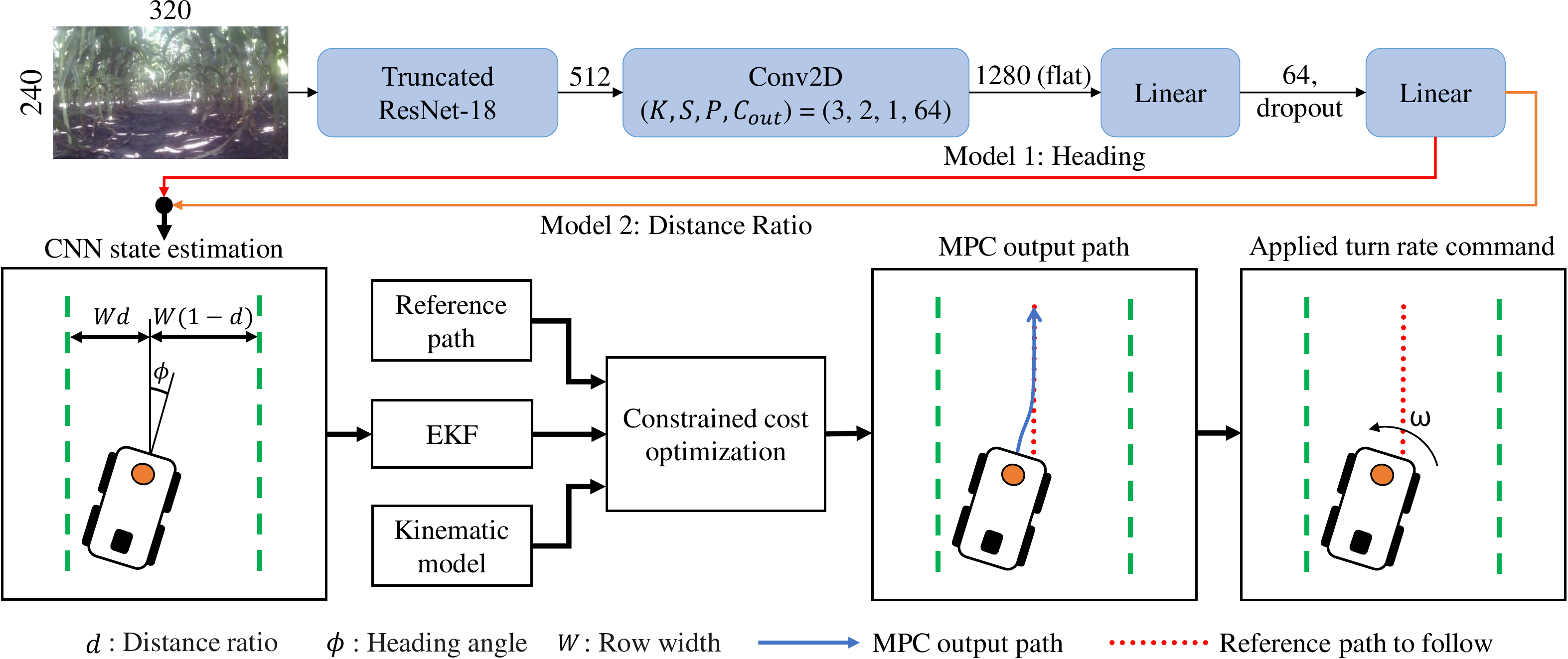}
   \caption{\textbf{\CF Overview}. We use a convolutional network to output robot heading and placement in row. This is used to compute the row center which is used as a reference trajectory. A model predictive controller converts reference trajectories to angular velocity commands.}
\figlabel{architecture}
\vspace{-0.2in}
\end{figure*}

\noindent \textbf{Autonomous Navigation in Agricultural Fields.}
GPS, alone and in combination with IMU and RTK corrections, is commonly used for outdoor navigation for tractors and over-canopy agricultural robots~\citep{reid2000agricultural, bak2004agricultural, bakker2011autonomous, zhang2017development,kayacan2018embedded, cordesses2000combine, kayacan2018high}.  Under-canopy navigation is concerned with autonomous row-following between the rows of crops. In such under-canopy environments, GPS suffers from significant multipath errors and signal attenuation under the canopy~\citep{higuti2019under}, furthermore RTK correction signals aren't always available.
As an alternative, \lidar data along with heuristics based algorithms for row-following have been used for under-canopy and orchard navigation~\citep{hiremath2014laser, barawid2007development, velasquez2020reactive, higuti2019under,Velasquez2021Multisensorfusion}. However, \lidar is costly, sensitivity to noise, and cannot sense semantic or contextual information.

This has motivated vision-based navigation systems. Past work in vision-based agricultural navigation can be classified into over the canopy~\citep{zhang1999agricultural,garcia2017automatic,zhai2016multi,jiang2015automatic,ball2016vision}, under-canopy in orchards~\citep{subramanian2006development,radcliffe2018machine, bergerman2015robot, aghi2020local} and under-canopy in row crops and horticultural crops~\citep{xue2012variable,gu2020path}. Vanishing lines based heuristics was commonly used in these works. In orchards and over-canopy visual navigation setting crop rows are clearly visible, which makes heuristic based line fitting possible. However, these algorithms do not directly apply to under-canopy navigation in commodity crops such as corn and soybean (the focus of this paper) where the row-spacing is much tighter (10$\times$ smaller than orchards), there is a high degree of visual clutter, complete and frequent occlusion of the camera by leaves, presence of weeds, crop residue on the ground, and changing visual appearance as the crop grows (see \figref{corn} and \figref{cvl_segmentation} for examples). Incidentally, corn and soybean acerage is atleast 10$\times$ larger than orchards. Recent visual servoing with RGB-D has been used for orchar navigation~\citep{aghi2020local}, however this approach will not work in corn-soybean canopies due to visual clutter and small-size of crops earlier in the growing season. 

\noindent \textbf{Classical Navigation.}
Navigation in classical mobile robotics~\citep{thrun2005probabilistic, siegwart2011introduction} follows a modular approach with perception (simultaneous localization and mapping (SLAM)), path planning relative to generated map, and trajectory tracking control. There are various successful SLAM techniques for this in structured and static environments such as in urban self-driving and indoor navigation. However, geometric reconstruction and localization in deformable and dynamic under-canopy agriculture environments is challenging.  
Furthermore, geometric approaches equate traversability with free space. While generally true, in off-road field settings this is not true (short weeds are fine to run over, hanging plant leaves can be run into), and necessitates the use of learning. Visual-inertial odometry (VIO) based approaches (\eg \citep{qin2018vins}) that are common in other outdoor navigation tasks are not useful here without a pre-built map, or GPS waypoints to close the loop and prevent drift (see \figref{rgbd_comparison}), or navigation in non-straight rows.

\noindent \textbf{Learned Navigation.}
Researchers have used machine learning for navigation and locomotion in situations where heuristics have failed. Learning has been used in different ways: \citep{gupta2017cognitive, zhu2017target, wijmans2019dd} learn high-level semantic cues and statistical regularities for navigation, \citep{kumar2018visual, 2020CorlEgo} use learning to provide robustness to actuation noise for path following, while \citep{gandhi2017learning, sadeghi2017cad2rl, pan2017agile, Ross-2013-7675, bojarski2016end} rely on learning to reduce or eliminate the dependence on expensive sensors for collision-free local navigation. Our work falls into this last category. Our use of learning not only eliminates dependence on \lidar, but surpasses its performance through better discrimination between traversible and intraversible areas by use of learning on camera images. Research in this last category can be further distinguished based on the policy design and supervision used for training.  Given the infeasibility of simulation, challenging terrain, lack of a reliable unsupervised self-supervision signal (as used in BADGR \citep{kahn2020badgr}), and difficulty of large-scale field experiments, renders  reinforcement learning, imitation learning, and self-supervision based methods infeasible for our task \citep{sadeghi2017cad2rl,pan2017agile,pan2017agile,gandhi2017learning, kahn2020badgr, qi2020learning, hadsell2009learning}. Also, lack of large-scale datasets for training has prevented the use of machine learning (over-canopy datasets \eg~\citep{chiu2020agriculture, pire2019rosario}, and urban self-driving datasets \eg~\citep{chang2019argoverse} exist, but aren't useful for under-canopy training). Therefore, we employ a modular approach \citep{bansal2019combining, muller2018driving, chaplot2020learning, meng2020scaling} and use supervised learning for training the perception module. Eliminating trial-and-error from learning improves sample efficiency, and the use of an analytical low-level controller allows easy generalization over varying terrains. Our contribution is in the design and experimental validation of a modular autonomy system in unique, challenging agricultural settings. 

\noindent \textbf{Learned Lane Following.}
Crop row following is similar to lane following in context of self-driving cars, however is much more challenging given no clear lane markings and extreme amounts of clutter. Past lane following works use reactive control based on traditional vanishing line estimation~\citep{zhang2012monocular}. However, vanishing line estimation is brittle. Consequently, recent works employ learning. \citep{chang2018deepvp} trained a vanishing point estimation network from an urban driving dataset with clearly visible lanes. Such lanes are not directly visible in our cluttered under-canopy environments (See \figref{cvl_segmentation} and \figref{diversity}). Thus, those models won’t work, as is, in our setting. Second, they only output the vanishing point which only tells us about the heading and not the distance ratio (see Appendix Section IV-E). Our method bypasses having to estimate the vanishing point and directly outputs all the necessary information required for the robot to navigate in under-canopy. Therefore, our method is the most direct and efficient way to achieve under-canopy row following. \citep{lyu2019road,neven2018towards, zou2019robust} predict semantic segmentation of the scene to estimate lane boundaries, while \citep{chen2017lanekeeping, simmons2019training, chi2017deep} employ end-to-end learning to directly output control commands via classification or regression). The former techniques require fine-grained pixel level annotations for training, and real-time inference is computationally expensive. End-to-end control is impractical in our setting as mentioned above. \citep{gurghian2016deeplanes} learns to predict the location of lane in the image to estimate distance but does not predict heading and distance directly. \citep{chen2015deepdriving} show that CNNs can be trained to predict driving affordances in uncluttered simulation environments where lane markings are clearly visible. In contrast, our work provides substantial experimental results that demonstrate that CNN based state estimators can lead to high-performing autonomous navigation systems capable of operating in the wild cluttered under-canopy fields, surpassing the current default practice of using a LiDAR. 

Closest to our work, Gu \etal~\citep{gu2020path} use learning to detect corn stalks and fit lines. This approach suffers when corn stalks are not visible, and has not been validated in real corn fields. We follow an implicit approach to directly estimate the states (row-relative heading and offset). This allows us to train a machine learning system that is robust to these challenges, as shown by our extensive in-field validation.

\section{System Design}
\seclabel{system}

\figref{architecture} shows an overview of our presented system. Images from on-board \rgb camera on the robot are processed through a convolutional network to predict robot heading $\phi$, and relative placement $d$ between crop rows. This relative placement is converted into the robot's distance from the left and the right crop rows by multiplying with the lane width.
These heading and distance predictions are filtered using a Bayesian filter (we use the Extended Kalman Filter) that optionally also fuses them with high-frequency input from an inertial measurement unit. The filtered heading and distances are used to generate a course correcting reference path in the robot coordinate frame. A model predictive controller is used to compute angular velocity commands to achieve this reference path. A lower-level proportional–integral–derivative (PID) controller is used to track the commanded angular velocity. 

In this section, we describe the robot platform, the CNN architecture, the Extended Kalman Filter, and the model predictive controller. We describe the data collection and ground truth generation procedure in \secref{dataset}.

\noindent \textbf{Robot Platform.} TerraSentia is an ultra-compact 4-wheeled skid-steering mobile robot designed to drive through fields and collect data. It has a Raspberry Pi 3 on-board for lower-level motor control and an Intel i7 NUC for data processing and navigation. Note that our unit had no discrete GPU, so the integrated Intel GPU is used for model inference. This robotic system is equipped with various sensors but only 4 are relevant to this paper. There is a dedicated GPS module that determines baseline autonomous driving performance (when GPS signal is reliable). The current \lidar-based autonomy is fueled by the 2D horizontal-scanning \lidar (Hokuyu UST-10LX) and a 6 DOF Inertial Measurement Unit (IMU). Finally, our approach utilizes only the forward facing, $720$p at $30$ fps monocular camera sensor (OV2710) and an IMU. We note specifically that since \lidar is not utilized in our presented visual system, no \textit{explicit} real-time depth signal is available to the model.

\noindent \textbf{Perception Model.}
We choose a learning approach due to its superior generalizability compared to color-based segmentation navigation proposed by previous works. \figref{cvl_segmentation} shows the classical system's failure to segment the lane in common late stage data. \CF's perception model takes in $320 \times 240$ \rgb images and outputs the robot heading (in degrees) and its relative placement in the crop row. \figref{lane} shows how the heading and the relative placement is defined. Heading $\phi$ is the angle of the robot relative to crop rows. The relative distance $d$ is the ratio of the distance to the left of the row to the lane width, \ie $d = \frac{d_L}{d_L + d_R}$, where $d_L$ and $d_R$ are the distances to left and right crop rows.

The perception model uses a ResNet-18~\citep{he2016deep} backbone that has been pretrained on ImageNet~\citep{deng2009imagenet}. We truncate ResNet-18 right before the average-pooling layer, and add in an additional convolutional layer, a fully connected layer, dropout, and final prediction layer. The final prediction layer outputs the heading $\phi$, and the distance ratio $d$. We found that independent networks to predict heading and distance ratio worked better than a single joint network.

\begin{figure}[t]
\centering
\insertWL{0.75}{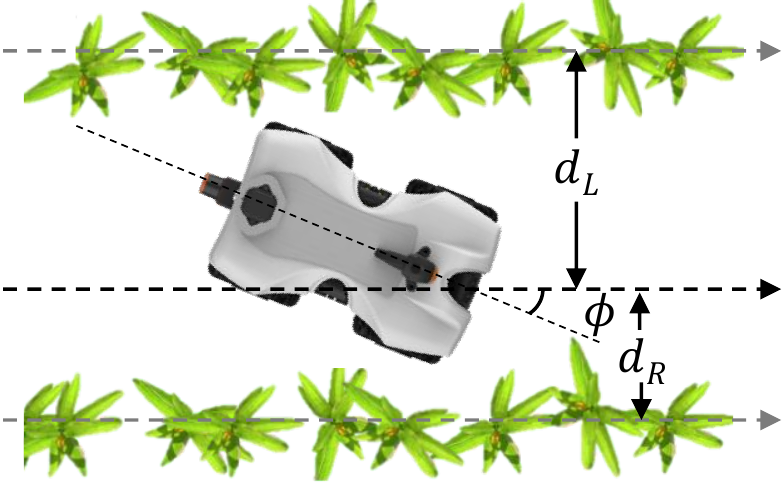}
\caption{Our method uses the robot's heading, $\phi$ and ratio of distance from the left and the right crop row, $d = d_L/(d_L+d_R)$, as the intermediate representation between perception and planning.}
\figlabel{lane}
\vspace{-0.2in}
\end{figure}

\noindent \textbf{IMU Fusion with Extended Kalman Filter.}
An Extended Kalman Filter \citep{chowdhary2013gps, farrell2008aided,siegwart2011introduction} was used to reduce the effect of uncertainties in distance and heading estimations by fusing the inertial data with the vision data. We used $s = \left(\begin{smallmatrix} d_L & d_R & \phi \end{smallmatrix}\right)^T$ as the state. State $s_k$ evolves over time as per the prediction function $f(s_{k-1}, u_{k-1})$ (derived using the robot's kinematics, see supplementary). Here $s_{k-1}$ is the state at the previous time step, and $u_{k-1}$ is the linear and angular velocity at the previous time step. Robot's linear speed $v$ and angular speed $\omega$ are calculated from wheel encoders, and IMU respectively. We assume additive zero-mean Gaussian process and measurement noise. As we directly observe $s$, the measurement function is an identity function. Output from the CNN is used in the update step. More details about the form of the prediction function, and co-variances of the Gaussian noise are provided in the supplementary material.

\noindent \textbf{Model Predictive Controller.}
We used a non-linear Model Predictive Controller (MPC) to generate angular speed commands to the robot given the reference path to be followed, as shown in \figref{architecture} \citep{kayacan2018embedded, kayacan2018high}. MPC uses the fused output states $s = \left(\begin{smallmatrix} d_L & d_R & \phi \end{smallmatrix}\right)^T$ from the EKF, the Unicycle kinematic model (see supplementary) of the robot and reference path, which is a straight line through the center of the lane, to solve a constrained optimization problem with the minimum and maximum curvature radius as the constraints. The output is a path defined in terms of the curvature $\rho$, which determines the angular velocity $\omega = \rho \; v$ where $v$ is the linear velocity. The angular speed for the first point in the output path is applied and the optimization process is repeated. A  PID controller is used to maintain the commanded angular speed, based on feedback from IMU's yaw angular speed.

\begin{figure}[t]
\centering
\includegraphics[width=1.0\linewidth]{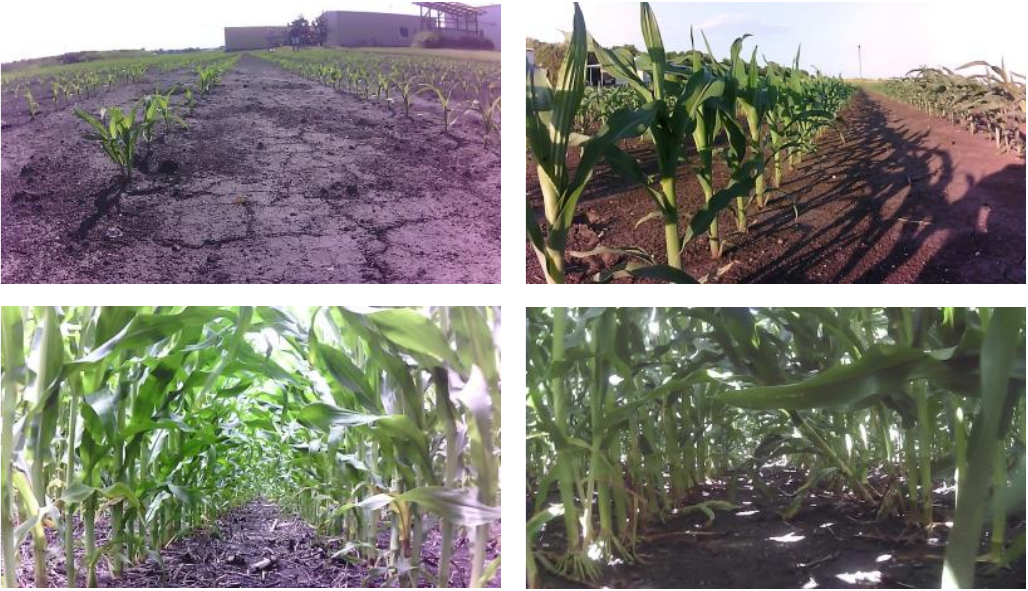}
\caption{Sample images from the collected dataset.}
\figlabel{corn}
\vspace{-0.2in}
\end{figure}

\section{Data Collection and Ground Truthing}
\seclabel{dataset}

Given lack of any under-canopy agriculture datasets, we collected a large dataset by driving the TerraSentia robot under the canopy. We manually operated the robot in $19$ corn and $4$ soybean fields across Illinois and Indiana, and collected time-series data from the front-facing \rgb camera, \lidar, and IMU. We collected $2.7$ hours of corn data and $1.2$ hours of soybean data, and made sure to collect data for different growth stages. We also included data where the robot was driven in a zigzag manner. This was done to expose the perception models to a broader distribution of data that may be experienced during autonomous runs. 
\figsref{corn}{soy} shows sample corn and soybean images from the dataset. We note the variability in appearance, occlusion, challenging illumination (shadows, low-light under the canopy), challenging terrain, and leafy plants. This raw data and a subset of annotations will be made available upon acceptance.

\begin{figure*}
\centering
\insertW{1}{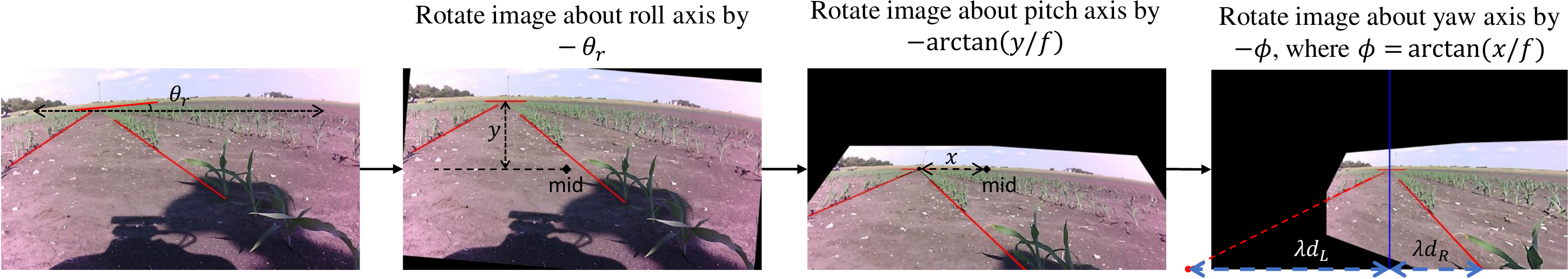}
\caption{\textbf{Ground truthing procedure.} Using the horizon annotations, we correct for the camera roll, and pitch. After this, heading, $\phi$ can be calculated by looking at the crop row vanishing point, and distance ratio can be computed from the intercepts of the crop row lines in the heading corrected image, as ${d_L}/({d_L+d_R})$.}
\figlabel{gt_flow}
\end{figure*}

\textbf{Ground Truthing.} To train our perception model from \secref{system}, we need labels for robot heading and the ratio of the distance from the left and the right crop rows. Preliminary investigation of using \lidar for extracting this information for training wasn't fruitful. Hence, we gathered human labels.

However, asking humans for such geometric labels is not easy. Unlike semantic labels, such metric geometric quantities are non-trivial for humans to label. As an example, consider images in \figref{corn}, and consider speculating the robot heading and placement in the row. 
To circumvent this issue, we designed an indirect annotation procedure. We asked humans to label the horizon and the vanishing lines corresponding to the crop row (\figref{annotations}~(left)). This together with the camera calibration information allows us to recover the robot heading and placement in row using projective geometry. \figref{gt_flow} provides an overview of the different steps involved in computing these quantities from the annotated images. For the case where the horizon is not visible, we instead ask humans to mark out vertical crop stalks (\figref{annotations}~(right)). This allows us to estimate the vanishing point for the vertical direction which readily provides the slope of the horizon. Precise formulae and derivations are provided in the supplementary material. 

We annotated a total of $25,296$ corn images. 28\% of these are from early growth stage, while 72\% are from late growth stage. We split the dataset into a training and a validation set (83\% training, 17\% validation). We made sure that data from the same video is either entirely in the training set, or entirely in the validation set. Our main experiments use this corn data. We also labeled $10,685$ soybean images (54\% early, 46\% mid) to study transfer across crops. 

\begin{figure}[t]
\centering
\insertWL{1.0}{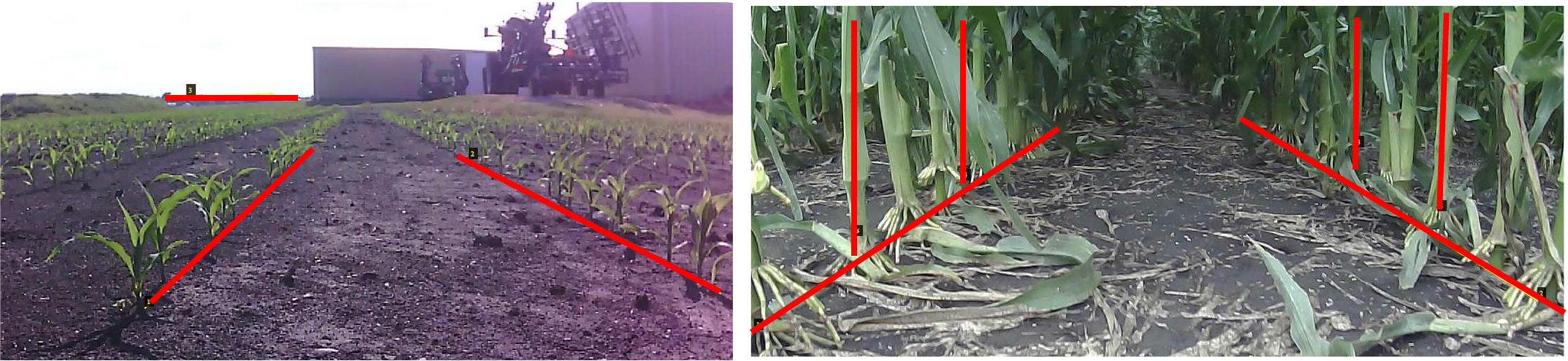}
\caption{\textbf{Annotations.} We annotate the horizon and crop rows for early season images~(left). For late season images when the horizon is not visible, we annotate the vertical corn stalks~(right).}
\figlabel{annotations}
\vspace{-0.15in}
\end{figure}

\section{Experimental Results}
Our experiments are designed to test the autonomous crop row traversal capability of our proposed system, effectiveness of the proposed modular policy, and data efficiency and generalization of our learned models. We evaluate these aspects through a combination of offline and online (field) experiments. Offline experiments are conducted on our collected dataset. They allow us to systematically study data efficiency and model generalization, and help us chose models for online experiments. Online experiments are conducted in the field, and allow us to study the interplay between perception and control systems. We also conduct end-to-end evaluation for the task of crop row traversal, and compare against an existing system based on \lidar \citep{Velasquez2021Multisensorfusion}.

\subsection{Offline Evaluation of Perception Model}
Offline evaluation of the perception module is conducted on the collected dataset. All experiments except ones for generalization across crops, use the corn to train and test.
\\ \noindent \textbf{Metrics.} We measure prediction performance using L1 error in heading and distance ratio predictions, $\phi$ and $d$.
\\ \noindent \textbf{Training.} We used ResNet-18~\citep{he2016deep} pretrained on ImageNet~\citep{deng2009imagenet} to initialize our models. Models were trained to minimize the $L2$ loss with the Adam optimizer~\citep{kingma2014adam} for 50 epochs. We started with an initial learning rate of $10^{-4}$ and dropped it by a factor of 10 at 40\textsuperscript{th} and 45\textsuperscript{th} epochs. All layers of the network were optimized.
\\ \noindent \textbf{Results.} \tableref{basic} presents the performance of our CNN models. We experimented with 2 variants: predicting heading and distance ratio separately using two models, and a single multi-task network. 
For reference, we also report the performance for a trivial predictor that always predicts the median heading and distance ratio from the training set. This measures the hardness of the task, puts performance of our model in context.

\setlength{\tabcolsep}{10pt}
\begin{table}
\resizebox{\linewidth}{!}{
\begin{tabular}{lS[table-format=1.2]cclS[table-format=1.2]l}
\toprule
\multirow{2}{*}{\textbf{Model}} & \multicolumn{2}{c}{\textbf{Mean}} & \multicolumn{2}{c}{\textbf{Median}}  & \multicolumn{2}{c}{\textbf{95\%ile}} \\ \cmidrule(lr){2-3} \cmidrule(lr){4-5} \cmidrule(lr){6-7} 
             & $\phi_{err}$ & $d_{err}$  & $\phi_{err}$  & $d_{err}$  & $\phi_{err}$  & $d_{err}$     \\ \midrule
\textbf{Baseline}    & 11.41     & 0.48      & 8.81      & 0.48      & 30.33     & 0.65 \\
\textbf{Combined}    & 2.24      & 0.08      & 1.39      & 0.06      & 5.37      & 0.20 \\
\textbf{Separate}    & \textbf{1.99}      & \textbf{0.04}      & \textbf{1.21}      & \textbf{0.03}      & \textbf{4.71}      & \textbf{0.10} \\ 
\bottomrule
\end{tabular}}
\caption{\textbf{Perception Module Performance:} We report L1 error in heading (in $^\circ$) and distance ratio prediction. The trivial baseline model always predicts median $\phi$, $d$ from the training set. The combined model learns heading and distance simultaneously, but ultimately performs worse than individually trained models.}
\tablelabel{basic}
\vspace{-0.2in}
\end{table}

Both models worked well, with the separate model variant working better. Our best model achieves an average L1 error of $1.99^\circ$ for heading, and $0.04$ for distance ratio. Inference speed for this model on the robot was around $20$ FPS, which is fast enough for accurate control (more on this in \secref{data-efficiency}). Our main field experiments are conducted with this model. 

\subsection{Comparison with Classical Baselines}
\begin{figure}[t]
\centering
\insertWL{1.0}{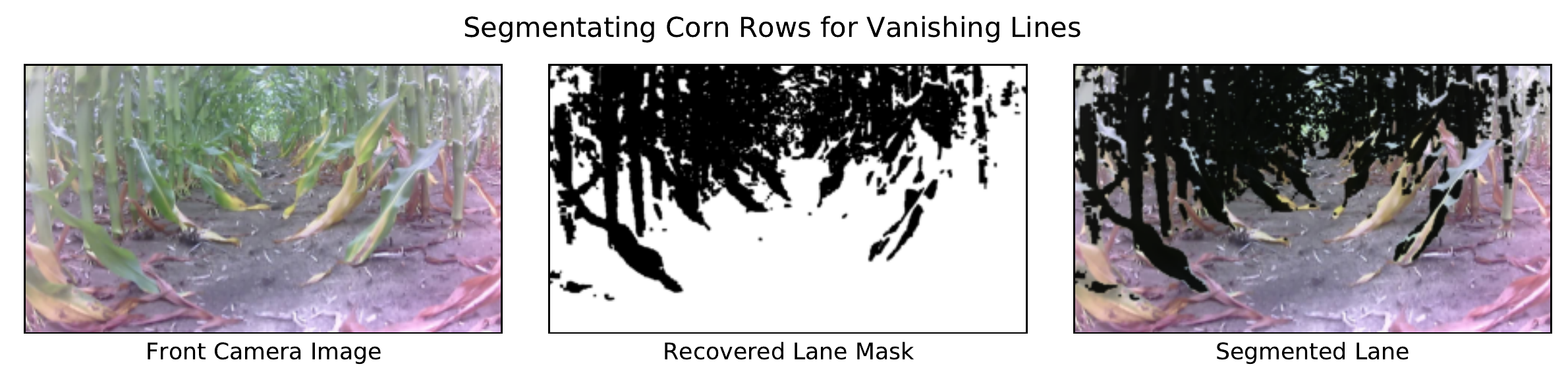}
\caption{Classical color-based vanishing line segmentation on late stage data according to related works \citep{bergerman2015robot, xue2012variable}. We see that crop segmentation does not produce a clear visual of the lane, so automatic vanishing line based lane-following is not possible. In particular, extraneous leaves artificially alter the boundaries of the lane.}
\figlabel{cvl_segmentation}
\vspace{-0.25in}
\end{figure}
\begin{figure}[t]
\centering
\insertWL{0.8}{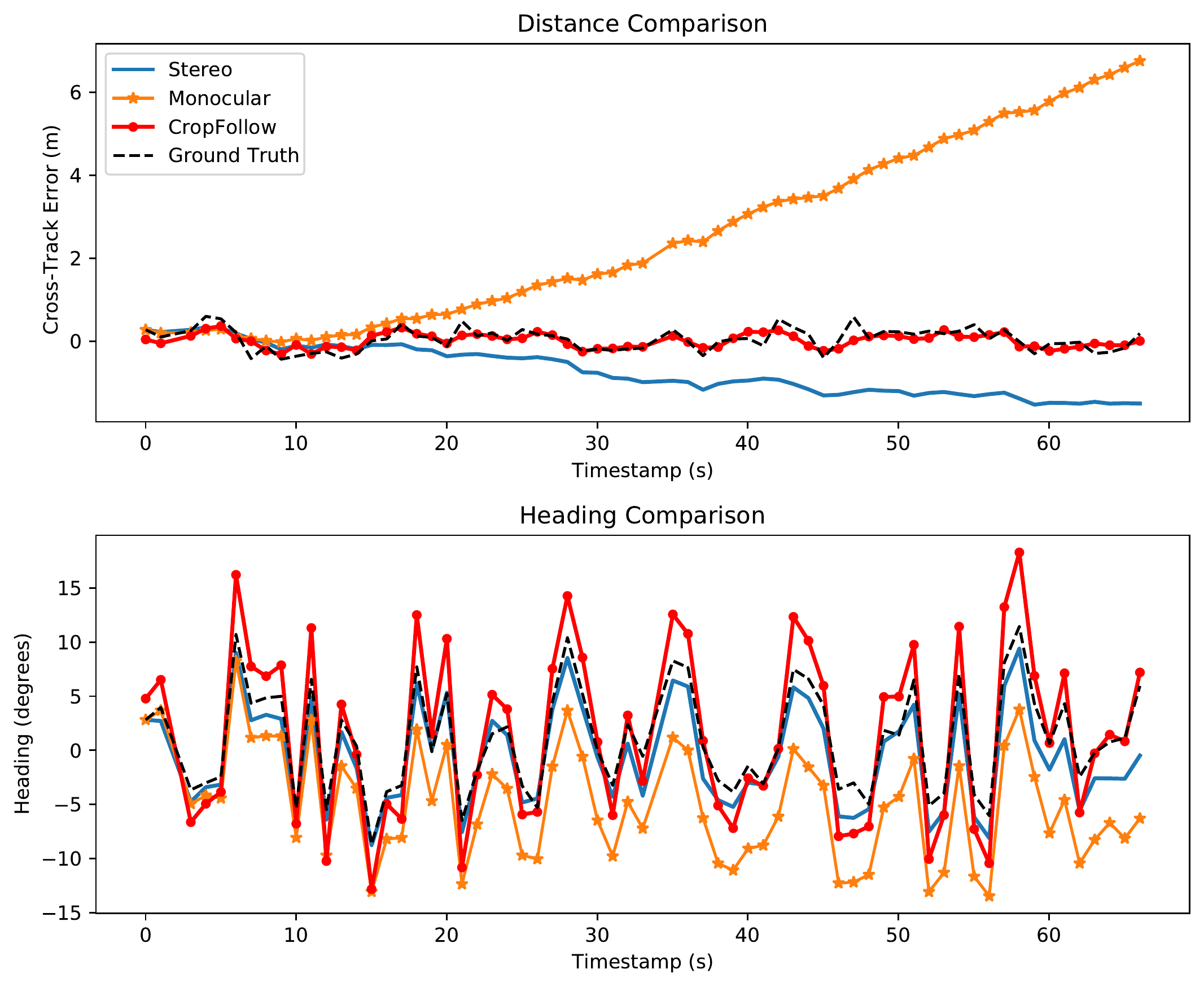}
\caption{\textbf{\CF \vs Vins-fusion mono \vs Vins-fusion stereo.} We compare the cross track error (CTE) (offset distance of the robot from the middle of the lane) and heading of \CF, Vins-fusion with mono and Vision-fusion with stereo IR at different frames in a trajectory. \CF shows better CTE than Vins-fusion.}
\figlabel{rgbd_comparison}
\end{figure}

 Color-based segmentation is a common first step in classical vanishing lines based row following literature. \figref{cvl_segmentation} shows the results of automatic color-based segmentation on common late stage data. We see that the segmented lane is not clear. This validates \CF's learning-based approach as a general navigation system for all growth stages across the season. To compare with a feature matching based VIO algorithm, Vins-fusion was used as the baseline~\citep{qin2018vins}. To compare with stereo based Vins-fusion as well, data collected from Intel Realsense D435i camera was used only for this experiment and recommended intrinsic values from Realsense library was used as Vins-fusion parameters. \figref{rgbd_comparison} demonstrates the heading and cross track error of \CF, Vins-fusion with monocular RGB camera and with stereo IR camera. Note that in case of distance the plot shows cross track error (offset distance from the middle of the lane) and not the relative error with respect to the ground truth. The ground truth was calculated by annotating vanishing lines and horizon (same approach as training labels for \CF). Ground truth heading and distance ratio at first frame was used to initialize Vins-fusion localization (both monocular RGB and stereo IR). \CF is vastly superior to Vins-fusion in distance prediction as seen from very similar cross track error as ground truth and is comparable in heading. Although Vins-fusion shows comparable heading tracking, it suffers significantly from position drift which is orders of magnitude greater than the lane width between crop rows (about 0.75m) making it impractical to use for row following. This is because there is no opportunity for loop closure in long crop rows. This validates reactive navigation as pursued in \CF is a valid approach for row following.

\subsection{In Field End-to-End System Evaluation}
\seclabel{main-eval}
We conducted end-to-end system evaluation with the model described above. We compared the performance of the following 2 systems, along with 2 variants each:
\begin{itemize}[noitemsep,leftmargin=*]
\item \textbf{\CF (w/ IMU)}. This is our proposed system that uses the above CNN model for heading and distance ratio prediction, EKF for fusing IMU information, and MPC for executing control commands. We also compare with a variant that does not use IMU information (denoted by \CF (w/o IMU)).
\item \textbf{\lidar System \citep{Velasquez2021Multisensorfusion} (w/ IMU)}. This system uses readings from the \lidar mounted on top of the robot to estimate the robot heading and distance from the crop rows using line fitting. Other parts of the system are same as our system: Use of an EKF to fuse information from the IMU, and use of MPC for generating control commands. We also compare to a variant that does not use IMU information (denoted by \lidar System~\citep{Velasquez2021Multisensorfusion} (w/o IMU)).
\end{itemize}

\begin{figure}
\centering
\includegraphics[width=1.0\linewidth]{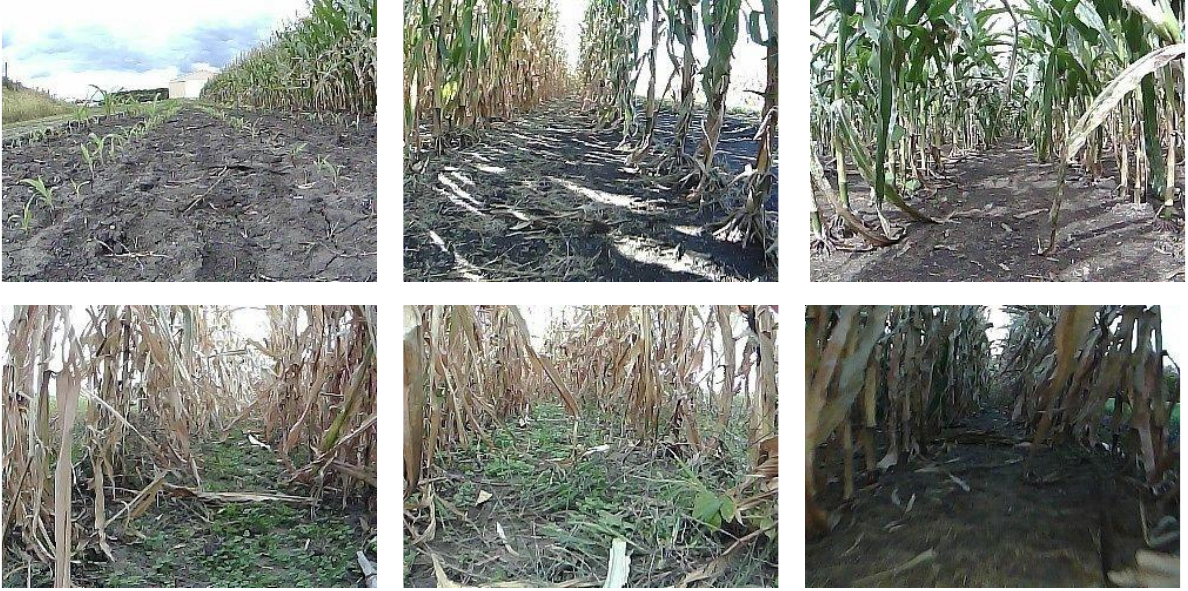}
\caption{Sample images from field trials. Bottom row consists of traditionally adverse conditions for vision-based navigation.} 
\figlabel{diversity}
\vspace{-0.2in}
\end{figure}

\noindent \textbf{Evaluation Methodology.} All 4 systems are tested on the same unique $4.85$ km. These $4.85$ km come from 15 different experiments that were done in different parts of the field, over different growth stages, different days, different time of the day, and weather conditions. While there is a lot of variability in these $4.85$ km, we attempted to minimized the variability in conditions for the 4 systems to ensure result comparability. Runs for the different systems for each of the 15 experiments were done one after another over the same routes, and with the same constant linear robot velocity of $0.6$ {m/s}. Run order for the different systems was randomized to prevent environmental bias. This experiment thus presents results pooled over field trials of $19.4$ km. For each method, we measure the number of human interventions needed to complete the experiment. Human interventions were required when the robot crashed into the corn stalks. This metric measures autonomy effectiveness.

\noindent \textbf{Results.} \tableref{interventions} reports the number of interventions for the 4 systems that we evaluated. We separately report results for early and late season experiments. Note that \lidar system from \citep{Velasquez2021Multisensorfusion} can't operate in early season data since early season corn stalks are shorter than the robot, and not detected by the 2-D \lidar. 
Our vision based systems works reasonably well. In late season when the \lidar based system does work, we note that it had more interventions than our system, 72 \vs 8 without IMU, and 13 \vs 7 with IMU. Thus, our presented vision-based system outperforms the \lidar based system, while also reducing sensing cost by 50$\times$ (\$30 for \rgb camera, while \$1500 for \lidar).
Note that these are paired experiments done over long run lengths ($4.85~$km), and the performance gap is statistically significant (with $p$-value $< 10^{-3}$). To further compare the without IMU versions of the \lidar system and \CF, we did an experiment where \lidar failed and \CF succeeded, and did 4 additional runs for each method (\tableref{repeatability}). We found \CF to work better than the \lidar system in all 5 trials. The quality of our output is further shown by the fact that our system is closing the loop only at about $20$Hz, \vs $40$Hz for the \lidar system, but still achieves a better end performance.

\setlength{\tabcolsep}{3pt}
\begin{table}[]
\resizebox{\linewidth}{!}{
\begin{tabular}{cccccc}
\toprule
\textbf{Growth} & \textbf{Length} & \textbf{\lidar} & \textbf{\lidar} & \textbf{\CF} & \textbf{\CF} \\
\textbf{Stage} & \textbf{(in m)} & \textbf{w/ IMU} & \textbf{w/o IMU} & \textbf{w/ IMU} & \textbf{w/o IMU} \\
\midrule
Early & 1120 & -     & -     & 3      & 4     \\ 
Late  & 3726 & 13    & 72    & 7      & 8     \\ \bottomrule
\end{tabular}}
\caption{\textbf{Field Experiments:} We report the number of interventions for the different methods. 
\lidar can't operate in early season as crops are too short. Our system can work under both conditions and requires interventions.}
\tablelabel{interventions}
\end{table}

\setlength{\tabcolsep}{8pt}
\begin{table}
\resizebox{\linewidth}{!}{
\begin{tabular}{lccccc}
\toprule
\textbf{Method} & \textbf{Trial 1} & \textbf{Trial 2} & \textbf{Trial 3} & \textbf{Trial 4} & \textbf{Trial 5} \\
\midrule
\lidar & 9 & 17 & 8 & 7 & 19 \\
\CF    & 0 & 0 & 0 & 0 & 0 \\
\bottomrule
\end{tabular}
}
\caption{We report the number of interventions of \lidar w/o IMU and \CF w/o IMU by repeating the test on the same row 5 times in the field. \CF outperformed \lidar in all trials. Variation in \lidar counts shows its sensitivity to noise.}
\tablelabel{repeatability}
\vspace{-0.2in}
\end{table}

\subsection{Training Data Efficiency and Generalization}
\seclabel{data-efficiency}
The above experiments demonstrate that our proposed system works. We next conduct experiments to measure data efficiency and generalization ability of our trained models. We investigate three questions: How much labeled data did we actually need to get good prediction and field performance? How much data do we need for the next crop? And what is the best use of annotation budget? We answer these questions through a combination of field and offline experiments. 

\begin{figure}[t]
\centering
\includegraphics[width=1\linewidth]{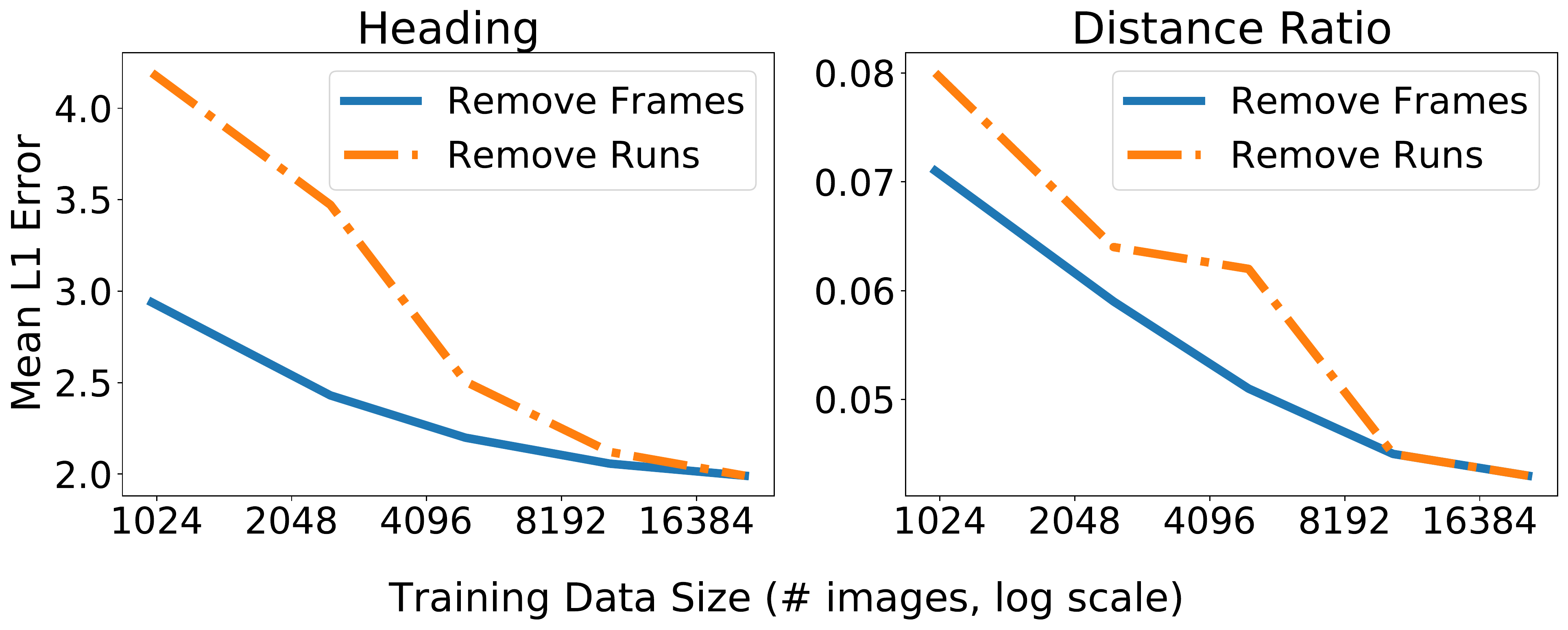}
\caption{Performance as a function of amount of training data. We sub-sample training data by either removing entire data collection runs, or by removing frames. Our perception model starts doing well even with small amounts of data.}
\figlabel{tds}
\end{figure}

\begin{figure}[t]
\centering
\includegraphics[width=0.8\linewidth]{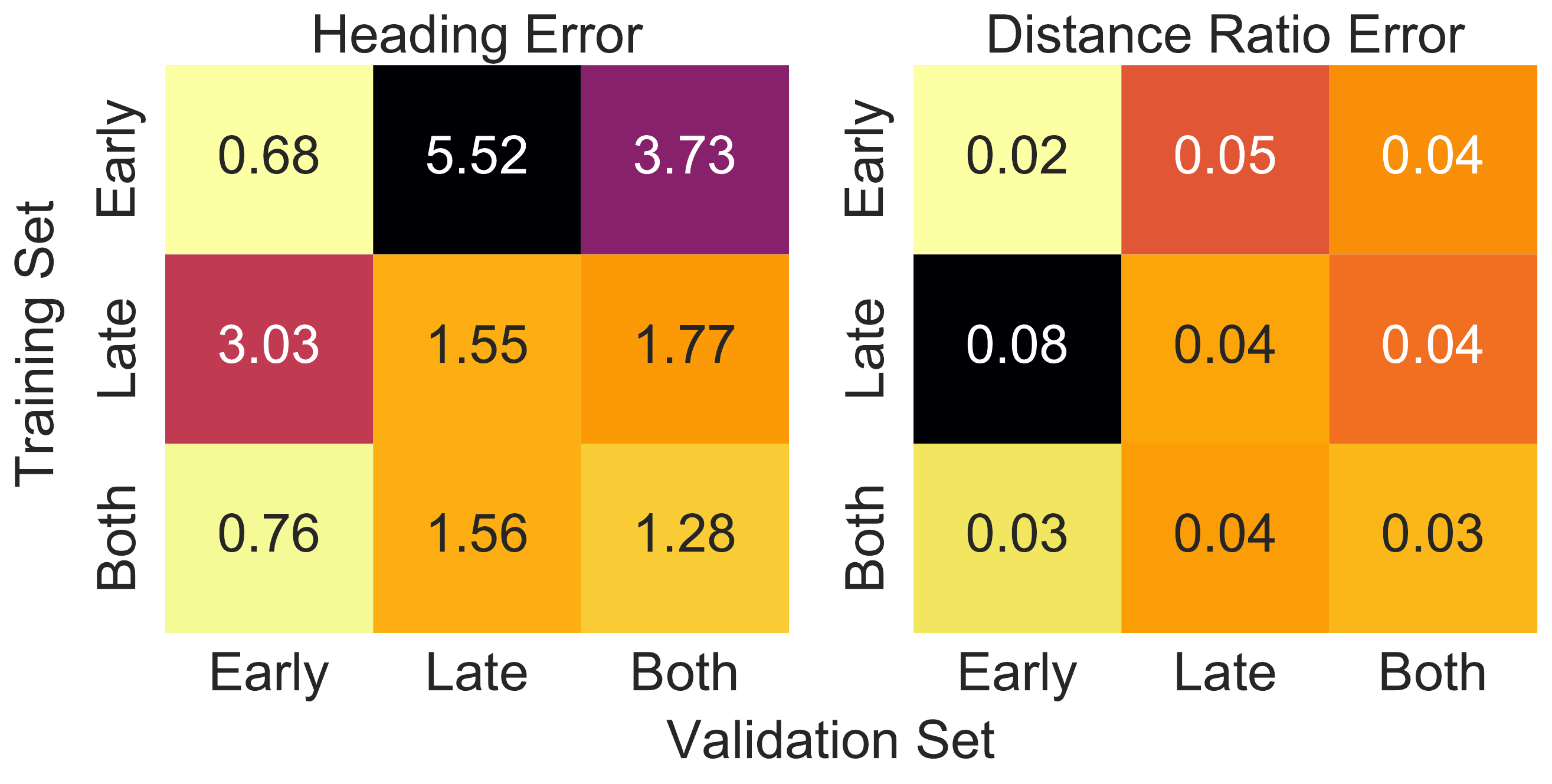}
\caption{Performance when training and testing on early \vs late \vs combined data. Models trained on only early or only late data don't generalize well, and training on combined data works best.}
\figlabel{growth_stage}
\vspace{-0.2in}
\end{figure}

\begin{table}
\resizebox{\linewidth}{!}{
\begin{tabular}{lccc}
\toprule
\textbf{Number of Training Images} & {100} & {1000} & {20986} \\ \midrule
\multicolumn{3}{l}{\textbf{Validation Metrics} (Mean L1 Error)} \\
\quad Heading Error       & {6.28}      & {4.19}       & {1.99}        \\ 
\quad Distance Error      & {0.09}      & {0.08}        & {0.04}        \\ \midrule
\multicolumn{3}{l}{\textbf{In field Metrics} (Number of Interventions)} \\
\quad \CF (w/ IMU) @ 22 FPS  & 0             & 0             & 0              \\ 
\quad \CF (w/ IMU) @ 10 FPS  & 0             & 0             & 0              \\ 
\quad \CF (w/ IMU) @ 5 FPS  & 4             & 0             & 0              \\ 
\quad \CF (w/ IMU) @ 2.3 FPS  & failed             & 0             & 0              \\ \midrule
\quad \CF (w/o IMU) @ 22 FPS & 0             & 1              & 0              \\ 
\quad \CF (w/o IMU) @ 10 FPS & 1             & 2              & 0              \\ 
\quad \CF (w/o IMU) @ 5 FPS & 2             & 0             & 0              \\ 
\quad \CF (w/o IMU) @ 2.3 FPS & failed             & 8             & 9              \\ \bottomrule
\end{tabular}}
\caption{Field and offline validation of models trained with 100, 1000 and 20986 images to study training data efficiency.}
\tablelabel{tds-field-tests}
\end{table}

\noindent \textbf{Data Efficiency for Corn.} 
We first measure the data efficiency of learning through offline experiments. We report the validation performance as a function of the amount of training data. We consider 2 versions obtained by sub-sampling a) at the level of data collection runs; b) at the level of frames. \figref{tds} plots performance as a function of training dataset size. We make two observations. First, models start performing well at around 10$K$ labeled images. Second, it is more beneficial to label images from many different runs, than many images from a few runs. 

We also study if we need data from all growth stages to learn a good model. \figref{growth_stage} reports validation performance on each growth stage for models trained on $6000$ images of either early stage, late stage, or an equal combination of both. We note that both models trained on a single growth stage have poor performance on the other growth stage. Our model that is trained on a blend of early and late stage data is most accurate throughout the entire season with an average error of $1.28^{\circ}$ and $0.03$ for heading and distance ratio respectively. 

\begin{figure}
\centering
\includegraphics[width=1.0\linewidth]{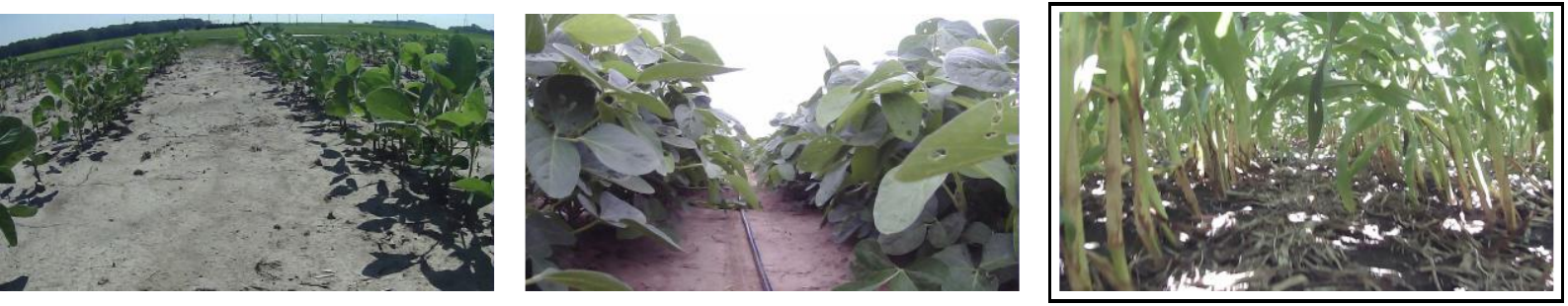}
\caption{Sample early and mid stage soybean. Note the stark difference to corn (right). Soybean is stouter with broader leaves.}
\figlabel{soy}
\end{figure}

\begin{figure}
\centering
\includegraphics[width=1.0\linewidth]{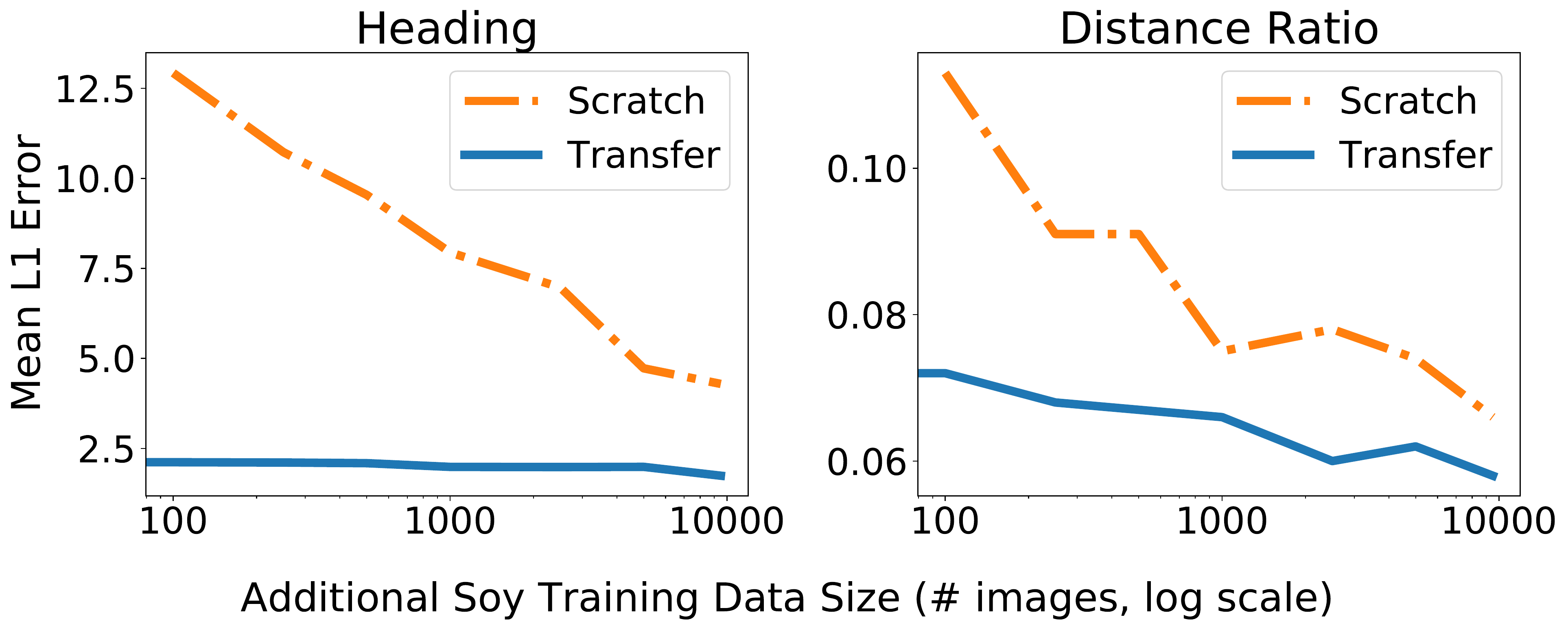}
\caption{\textbf{Generalization from corn to soybean.} Model trained on corn (\emph{Transfer}) generalizes well to soybean in comparison to training from ImageNet initialization (\emph{Scratch}).}
\figlabel{soy_generalization}
\vspace{-0.2in}
\end{figure}

\begin{figure*}[tbh]
\centering
\includegraphics[width=\linewidth]{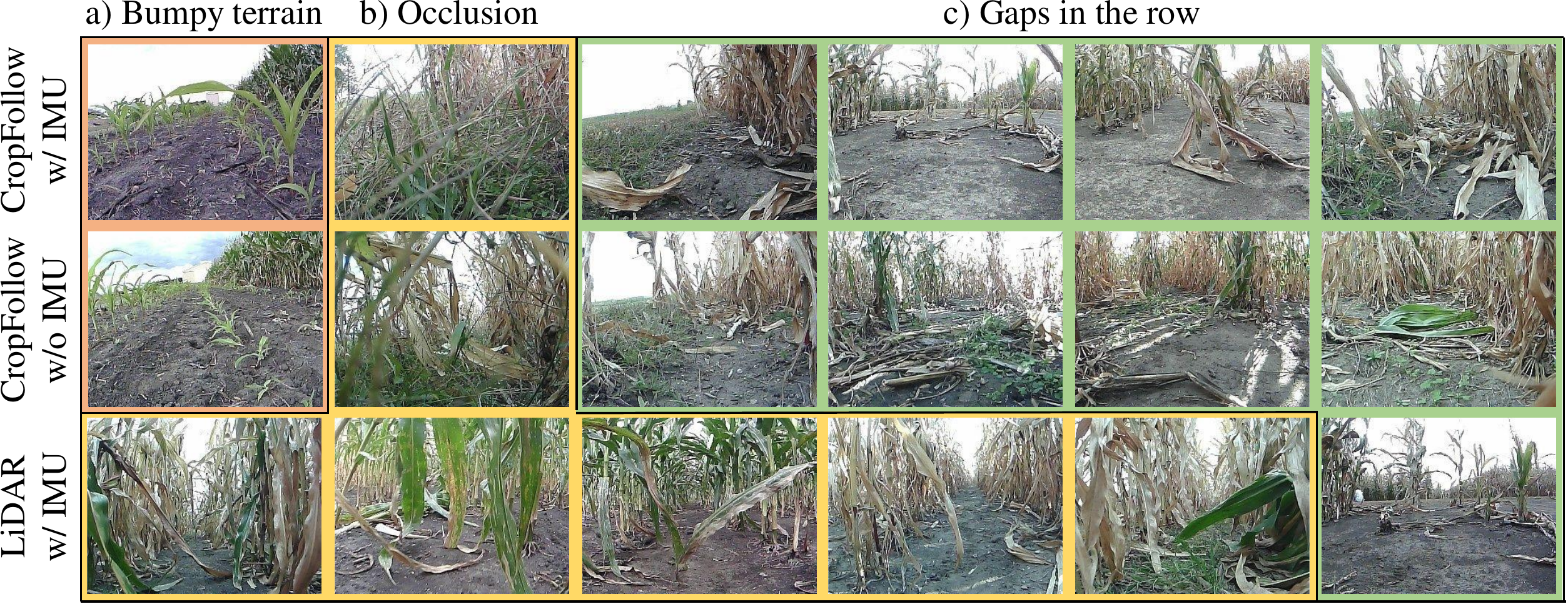}
\caption{Failure scenarios for the different navigation systems. We group them into modes: a) bumpy terrain causing noisy, blurry images, b) sensor occlusion from leaves, and c) gaps in crop rows.}
\figlabel{failures}
\vspace{-0.2in}
\end{figure*}

\noindent \textbf{Field Experiments for Data Efficiency for Corn.} 
However, note this is only performance of the perception module in isolation. It will be more instructive to look at the field performance of the whole system as a function of the training set size. \tableref{tds-field-tests} reports field performance of 3 models trained with $100$, $1000$ and $21K$ images (we took the models that sub-sampled data at the level of runs as they had a sharper drop in performance), in the same crop row of length $428$m. Interestingly, we note that at the base control frequency of $20$Hz, systems trained with as little as 100 images worked without interventions! It should be noted that this does not mean that 100 images are sufficient for robust and repeatable performance, but shows that the system learns quite a bit with little data, and the modular approach which leverages the IMU and a robust controller is capable of tolerating a less perfect perception system. Indeed, difference in performance is more evident at lower update rates. Perception models trained on larger datasets are likely more robust to extreme viewpoints and hence can recover better from off-center locations that may arise at lower update rates. These results provide information on allowable heading and distance ratio prediction error at different 
speeds and update rates with which amount of training data needed for training in new crops can be determined. It can be seen that with higher prediction errors, using IMU and higher model update rate makes the system robust.

\noindent \textbf{Generalization to Another Crop.}
We also study the data efficiency for enabling autonomous navigation for a new crop. We do this via offline experiments and measure how much additional training data is needed to adapt a model trained on corn to achieve good performance. \figref{soy_generalization} plots the validation metrics as a function of number of Soybean training images (\textit{Glycine max}, \figref{soy}), for our transferred model, and for a baseline model that starts from ImageNet initialization. We note strong transfer of the model trained on Corn. Even without any training on Soybean, our two models achieve good performance with a average error of just $2.20^{\circ}$ and $0.07$ for heading and distance. Although only from Corn to Soybean, this is a very desirable result. It suggests that our Corn model might already work in Soybean rows with minimal additional labeling. We leave field trials to future work.

\subsection{Error Modes and Stress Testing}
To understand the common error modes in our \CF and \lidar system, we visualized the front camera video stream before failures in field experiments. Also, to stress test the proposed \CF, field tests were conducted in a challenging field with a sharp curve, occlusions and gaps and experiments were also conducted to test the performance of \CF at increased speeds.
    
\noindent\textbf{Visualization of different error modes.}
\figref{failures} shows the different error modes in \CF and \lidar navigation system. Large gaps in crop rows was the common cause of failure in \CF (our training data did not include such cases). Sensor occlusion and bumpy terrain were the other rare causes of failures. In contrast, failure due to gaps was rarely observed in \lidar since it was specifically engineered to be robust to it. But because of its high sensitivity to noise, even minor sensor occlusion by leaves affects \lidar performance and leads to interventions. \CF's performance in gaps could be improved with adding training data whereas \lidar's occlusion problem is a sensor limitation.

\noindent\textbf{Stress testing.} To test the performance in challenging conditions, \CF (w/ and w/o IMU) was tested in a field with sharp curves, gaps and occlusion from weeds. 3 and 6 interventions w/ and w/o IMU respectively was observed in a test of 600m. Last row in \figref{diversity} shows the challenging condition in this field. Also, \CF's performance at higher speeds was tested. \CF showed same stable behavior at $1 m/s$ but oscillations in trajectory due to latency was observed at $1.4 m/s$ or more.

\section{Conclusion}
\label{sec:conclusion}
We presented a vision based autonomous under-canopy navigation system. Through a modular architecture and a learning-based approach we showed that machine vision can be applied for reliable and robust navigation in cluttered, changing, and harsh under-canopy environments. 25~km of real-world validation on an under-canopy robot demonstrated that our visual navigation approach is not only 50$\times$ more cost-effective than \lidar but also leads to fewer interventions. Our system forms a new benchmark for visual navigation under the canopy, and our openly accessible dataset (1030 labeled images and 24266 unlabeled images of our corn data) will enable further research. We hope our results and dataset pave the way for wider adoption of learned visual navigation systems in challenging application domains, such as agriculture and off-road driving. 

\section{Acknowledgments}
This paper was supported in part by NSF STTR \#1820332, USDA/NSF CPS project \#2018-67007-28379, USDA/NSF AIFARMS National AI Institute USDA \#020-67021-32799/project accession no.~1024178, NSF IIS \#2007035, and DARPA Machine Common Sense. We thank Earthsense Inc. for the robots used in this work and we thank the Department of Agricultural and Biological Engineering and Center for Digital Agriculture (CDA) at UIUC for the Illinois Autonomous Farm (IAF) facility used for data collection and field validation of \CF. We thank Vitor Akihiro H. Higuti and Sri Theja Vuppala for their help in integration of \CF on the robot and field validation.

\bibliographystyle{plainnat}
\bibliography{rss}

\twocolumn[{%
 \centering
 \LARGE Supplementary Material\\[1.5em]
}]

\section{Video}
Please see the accompanying video that provides an overview of our paper, and shows video executions of our robots. Video is encoded via H-264 MPEG4 and was tested to play well on Windows and MacOS through all regular media players such as Movies \& TV (Windows), QuickTime, VLC, and Google Chrome.

\section{IMU Fusion with Extended Kalman Filter}

An Extended Kalman Filter was used to reduce the effect of uncertainties in distance and heading estimations. The state vector was defined as $s = \left(\begin{smallmatrix} d_L & d_R & \phi & \omega \end{smallmatrix}\right)^T$ where $\phi$ is the robot's heading, $\omega$ is the angular velocity of the robot  and $d_R$  and $d_L$ are the distances to right and left rows respectively.  The process was modeled using \eqref{ekf} where actual state $s[k]$ was defined as a function of $f(\cdot)$ (shown in \eqref{stateFunction1}), the control inputs $u_k$ and the previous state $s[k-1]$, 
\begin{equation}
    \eqlabel{ekf}
    \begin{array}{ccl}
        s[k] & = & f(s[k-1], u[k]) + \omega_k \\
        z[k] & = & s[k] + \nu_k
    \end{array}
\end{equation}
here, $f(s[k-1], u[k])$ is defined as,
\begin{equation}
    = \left[
        \begin{array}{ccl}
            d_L [k]  \\
            d_R [k]  \\
            \phi[k] \\
            \omega[k]
        \end{array}
    \right]
    =
    \left[
        \begin{array}{ccl}
            d_L [k-1] - v \sin(\phi[k-1])\Delta t  \\
            d_R [k-1] + v \sin(\phi[k-1])\Delta t  \\
            \phi[k-1] + \omega \Delta t \\
            \omega[k]
        \end{array}
    \right].
    \eqlabel{stateFunction1}
\end{equation}

Both process noise $\omega_k$ and measurement noise $\nu_k$ were defined as zero mean Gaussian noises and their covariances  are $\left[\begin{smallmatrix} 0.001 & 0.001 & 0.01 & 0.01 \end{smallmatrix}\right]$ and $\left[\begin{smallmatrix} 0.05 & 0.05 & 0.05 & 0.5 \end{smallmatrix}\right]$ respectively, corresponding to the states and measurements $d_L$, $d_R$, $\phi$, and $\omega$. Those values are in the covariance matrices $\mathbf{Q}$ (for $\omega_k$) and $\mathbf{R}$ (for $\nu_k$).

The robot’s linear $v$ and angular $\omega$ velocities are used to estimate the states (\eqref{stateFunction1}) in the prediction step. $v$ is calculated from encoders and $\omega$ is obtained from IMU. In the update step, innovation occurs by considering the calculated values of $d_L$, $d_R$ and $\phi$ from 2 CNN networks. As the output of the distance CNN network is a distance ratio, it is necessary to convert it to a metric value by multiplying it with average lane width.

\section{Model Predictive Control}

The kinematic differential model is formulated for a skid-steering mobile robot as presented in Eq. \ref{eq:skid-model}.

\begin{equation}\label{eq:skid-model}
    \begin{aligned}
        \dot{x} &= v \; cos(\phi) \\
        \dot{y} &= v \; sin(\phi) \\
        \dot{\phi} &= \omega
    \end{aligned}
\end{equation}

The robot's states $(x, y, \phi)$ denote its bi-dimensional position and yaw angle, while inputs $v$ and $\omega$ denote its linear and angular speeds. Then, it's possible to transform the differential model to a discrete model and solve it using Euler's integration, as given by Eq. \ref{eq:discrete}.

\begin{equation}\label{eq:discrete}
    \begin{aligned}
        x[k] &= x[k-1] + v \; cos(\phi) \; \Delta t \\
        y[k] &= y[k-1] + v \; sin(\phi) \; \Delta t \\
        \phi[k] &= \phi[k-1] + \omega \; \Delta t
    \end{aligned}
\end{equation}

A transformation of $v \ \Delta t = \Delta s$ is adopted, and therefore $\omega \ \Delta t = \rho \ \Delta s$, where $\rho$ is the robot's instantaneous curvature. The new non-linear system is then given by Eq. \ref{eq:discrete-ds}.

\begin{equation}\label{eq:discrete-ds}
    \begin{aligned}
        x[k] &= x[k-1] + cos(\phi) \; \Delta s \\
        y[k] &= y[k-1] + sin(\phi) \; \Delta s \\
        \phi[k] &= \phi[k-1] + \rho \; \Delta s
    \end{aligned}
\end{equation}

The non-linear model is used as dynamic model of the process to predict future states, and a cost function over the receding horizon is the optimization cost function using as control input the curvature $\rho$.

\begin{equation} \label{eq:cost-function}
    \begin{aligned}
        \min_{\rho_i} \Bigg\{ &\sum _{i=1}^{N}w_{d_{e,i}}d_{e,i}^{2} + \sum _{i=1}^{N}w_{\phi_{i}}\phi_{error,i}^{2} + \\ 
        &\sum _{i=1}^{N-2}w_{\Delta \rho_{i}}({\rho_{i} - \rho_{i-1}})^{2} \Bigg\}
    \end{aligned}
\end{equation}

Where $d_{e,i}$ is the cross-track error, $\phi_{error,i}$ is the heading error, $\rho_{i}$ is the curvature command, $w_{d_{e,i}}$ is the weighting coefficient reflecting the relative importance of $d_{e,i}$, $w_{\phi_{i}}$ is the weighting coefficient reflecting the relative importance of  $\rho_{i}$, and $w_{\Delta\rho_{i}}$ is the weighting coefficient penalizing relative big changes in curvature.

The variables used in the Eq. \ref{eq:cost-function} are calculated as shown in Eq. \ref{eq:error-equations}, which shows the calculation of the cross-track error using geometry and the heading error as the difference between current robot's heading (equal to zero in the local robot's frame) and the desired path's heading. The optimization problem is subject to a single constraint inequality acting in the curvature command, which must lay in $-1/R_{max} \leq \rho_i \leq 1/R_{max}$. The $R_{max}$ is the maximum permissible curve radius the robot can follow, which is a tunable parameter for compromise between aggressiveness and to avoid robot to get stuck on difficult terrains.

\begin{equation} \label{eq:error-equations}
    \begin{aligned}
        \phi_{wp_i} &= \arctan2(wp_{y_i}-wp_{y_{i-1}}, wp_{x_i}-wp_{x_{i-1}}) \\
        R_U &= \sqrt{(wp_{x_{i-1}}-x_i)^2 + (wp_{y_{i-1}}-y_i)^2} \\
        \phi_U &= \arctan2(y_i-wp_{y_{i-1}}, x_i-wp_{x_{i-1}}) \\
        d_{e,i} &= R_U \; \sin(\phi_{wp} - \phi_U) \\
        \phi_{error,i} &= \arctan2(wp_{y_i}-wp_{y_{i-1}}, wp_{x_{i}}-wp_{x_{i-1}})
    \end{aligned}
\end{equation}

In Equation \ref{eq:error-equations}, $wp_{y_i}$ and $wp_{x_{i}}$ are the coordinates of the ith waypoint used as input in the MPC horizon. These waypoints are generated as a straight line that represents the middle of the crop row, whre this line is calculated from the distance ratio $d$, lane width $W$, and the estimated heading $\phi$ estimated from the vision algorithm and EKF. For each iteration, the cross-track error $d_{e,i}$ and the heading error $\phi_{error,i}$ are calculated  as shown in Equation \ref{eq:error-equations}, and they are used as functions for the minimization in Equation \ref{eq:cost-function}.

During the experiments, the parameters were used as described in Equation \ref{eq:parameters}.

\begin{equation} \label{eq:parameters}
    \begin{aligned}
        R_{min} &= 0.7 \\
        \Delta s &= 0.2 \\
        w_{d_{e,i}} &= \left\{ 
            \begin{array}{ll}
                120 & i=1,2,\dots,19 \\
                1200 & i=20 \\
            \end{array} \right. \\
        w_{\phi_{i}} &= \left\{ 
            \begin{array}{ll}
                100 & i=1,2,\dots,19 \\
                1000 & i=20 \\
            \end{array} \right. \\
        w_{\Delta \rho_{i}} &= 1000 \quad i=1,2,\dots,20 \\
    \end{aligned}
\end{equation}

\begin{figure*}[t]
     \begin{center}
     \includegraphics[width=1\linewidth]{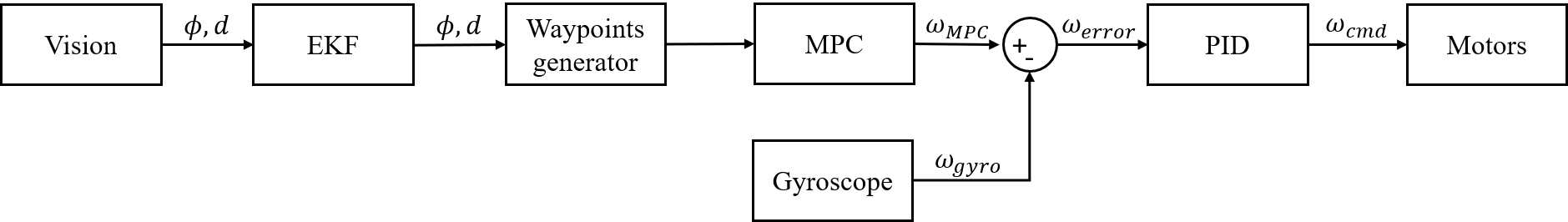}
     \end{center}
      \caption{Overall control diagram.}
     \label{fig:control-diagram}
\end{figure*}

A  PID controller is used as low-level controller to guarantee the predicted control effort is followed by the robot. The low-level controller uses the IMU's yaw angular speed as feedback to follow the angular speed command that comes from the MPC controller. As input to the MPC controller, a waypoints generator algorithm is used. The generated path is always straight and built in relation to the robot's local frame. The distance and angle of the straight line depends on the measured distance error and heading error from desired path. Figure \ref{fig:control-diagram} shows the overall control diagram, where $\omega_{MPC}$ is the yaw angular speed calculated from MPC algorithm, $\omega_{gyro}$ is the robot's yaw angular speed measured using the IMU's gyroscope, $\omega_{error}$ is the difference between the $\omega_{MPC}$ and $\omega_{gyro}$, and $\omega_{cmd}$ is the commanded yaw angular speed that is sent to the motors.

\section{Ground Truthing}
We use projective geometry to obtain the heading $\phi$ and distance ratio $d$ from our obtained annotations. We show the derivations for the different steps in this section. We assume a pinhole camera model, and assume camera's focal length to be $f$. 
We denote world coordinates with a capital letters $(X,Y,Z)$, and denote their projection in the image as $(x,y)$. Note under the pin hole camera model, $x = \frac{fX}{Z}$ and $y = \frac{fY}{Z}$. The $X$-axis goes right from the image center, $Y$-axis goes down from the image center, and the $Z$ axis goes into the scene from the camera center.

As noted our ground truthing process has 4 steps: camera roll correction (\secref{roll}), camera pitch correction (\secref{pitch}), heading estimation (\secref{heading}) and distance ratio estimation (\secref{distance}). We do these on top of annotated horizon and crop row vanishing lines. For early season data, we can directly annotate these. For late season data, the horizon is not directly visible, and we mark the corn stalks to estimate the horizon from the vanishing point of the vertical lines (\secref{horizon}). Our annotation procedure assumes: a) ground is flat, b) corn has been planted in parallel rows, c) corn stalks are vertical. We found these to be reasonable assumptions for the data that we were working with. 

Images are rotated about the camera center by the obtained roll, pitch and heading angles incrementally between steps, using homography $H = KRK^{-1}$, where $K$ is the camera matrix, and $R$ is the desired rotation about the camera center.

\subsection{Horizon Estimation}
\seclabel{horizon}
Let's assume the vertical stem lines in the world are in the direction $(D_X, D_Y, D_Z)$. Vanishing point of these lines is can be obtained by considering a line in this direction through a point $(A_X, A_Y, A_Z)$, projecting points on this line onto the image plane, and then taking the limit as the points tend to infinity,
\begin{equation}
    \eqlabel{horizon_estimation}
    \begin{aligned}
        &\lim_{\lambda\to\infty}\left(f_X\frac{A_X+\lambda D_X}{A_Z + \lambda D_Z}, f_Y\frac{A_Y+\lambda D_Y}{A_Z + \lambda D_Z}\right) \\
        = &\left(f\frac{D_X}{D_Z} , f\frac{D_Y}{D_Z}\right) = \left(v_x, v_y\right) \\
    \end{aligned}
\end{equation}
where $(V_X, V_Y)$ is the vanishing point. \\
Horizontal plane is given by the following equation,
\begin{equation}
    \eqlabel{horizon_plane}
    \begin{aligned}
        D_X X + D_Y Y + D_Z Z = c \\
        D_X\frac{fX}{Z} x + D_Y\frac{fY}{Z} + fD_Z = \frac{fc}{Z} \\
        D_X \cdot x + D_Y \cdot y + fD_Z = \frac{fc}{Z} \\
    \end{aligned}
\end{equation}
The horizon occurs when we go to $\infty$ on this plane, \ie $\lim_{Z\to\infty}$, \\
\begin{equation}
    \eqlabel{horizon_eqn}
    \begin{aligned}
        D_X \cdot x + D_Y \cdot y + fD_Z = 0 \\
        \frac{D_X}{D_Z} \cdot x + \frac{D_Y}{D_Z} \cdot y + f = 0 \\
    \end{aligned}
\end{equation}
We can substitute for $\frac{D_X}{D_Z}$ and $\frac{D_Y}{D_Z}$ from \eqref{horizon_estimation}, to obtain the equation of the horizon as,
\begin{equation}
    \eqlabel{horizon_eqn}
    \begin{aligned}
        v_x \cdot x + v_y \cdot y + f^2 = 0
    \end{aligned}
\end{equation}

Vanishing points for the vertical lines can be found by finding the point of intersection of the lines using least squares.

\subsection{Roll Estimation}
\seclabel{roll}
Suppose the camera is pitched down by pitch $\theta$ and has roll $\alpha$, then the surface normal of the ground plane is given as $(\sin\alpha\cos\theta, \cos\alpha\cos\theta, -\sin\theta)$. The equation of the ground plane is given by:
\begin{equation}
    \begin{aligned}
        X\cos\theta\sin\alpha + Y\cos\alpha\cos\theta - Z\sin\theta &= c \\ 
        \Rightarrow \frac{fX}{Z}\cos\theta\sin\alpha + \frac{fY}{Z}\cos\alpha\cos\theta - f\sin\theta &= \frac{fc}{Z} \\
        \Rightarrow x\cos\theta\sin\alpha + y\cos\alpha\cos\theta - f\sin\theta&= \frac{fc}{Z} \\
    \end{aligned}
\end{equation}
where in the last step, we have substituted image coordinates, $x$ for $\frac{fX}{Z}$ and $y$ for $\frac{fY}{Z}$, using the pinhole camera model. We get the equation of the horizon in the image plane, by seeing what happens as $Z \rightarrow \infty$:
\begin{equation}
x\cos\theta\sin\alpha + y\cos\alpha\cos\theta - f\sin\theta= 0
\end{equation}

We can compute the camera roll $\alpha$ from the slope of the horizon $h'$ in the image,
\begin{equation}
    \eqlabel{roll}
    \begin{aligned}
        h' &= -\frac{\cos\alpha\cos\theta}{\sin\alpha\cos\theta} = -1/\tan\alpha\\
        \Rightarrow \alpha &= -\arctan(1/h')
    \end{aligned}
\end{equation}

\subsection{Pitch Estimation}
\seclabel{pitch}
Assuming the image and all annotation lines have been corrected for roll, we notice that the ground plane will have a normal vector $(0, \cos\theta, -\sin\theta)$. Following a similar procedure as in \secref{roll}, we obtain the equation of horizon as: 
\begin{equation}
    \eqlabel{pitch_ground_points}
    \begin{aligned}
        0\cdot X + Y\cos\theta - Z\sin\theta &= c \\
        \Rightarrow 0 \cdot \frac{fX}{Z} + \frac{fY}{Z}\cos\theta - f\sin\theta &= \frac{cf}{Z} \\
        \Rightarrow 0 \cdot x + y\cos\theta - f\sin\theta &= \frac{cf}{Z} \\
    \end{aligned}
\end{equation}

As we tend $Z \rightarrow \infty$, 
\begin{equation}
    \begin{aligned}
        0 \cdot x + y\cos\theta - f\sin\theta &= 0 \\
        \Rightarrow y = f \tan \theta
    \end{aligned}
\end{equation}
Thus, $\theta$ can be found using the y-intercept of the horizon in the image, as $\arctan(y_\text{horizon} / f)$, where $y_\text{horizon}$ is the y-coordinate of the horizontal horizon line. 

\subsection{Heading Estimation}
\seclabel{heading}
Now that the image and annotation lines have been corrected for pitch and roll, heading $\phi$ can be obtained from the vanishing point of the crop row lines. The left crop row is in the direction of $(\sin \phi, 0, \cos\phi)$, and let us assume that it passes through the point $(A^l_X, A^l_Y, A^l_Z)$. A point on the the left crop row is given by: 

\begin{figure}[t]
\centering
\insertWL{0.9}{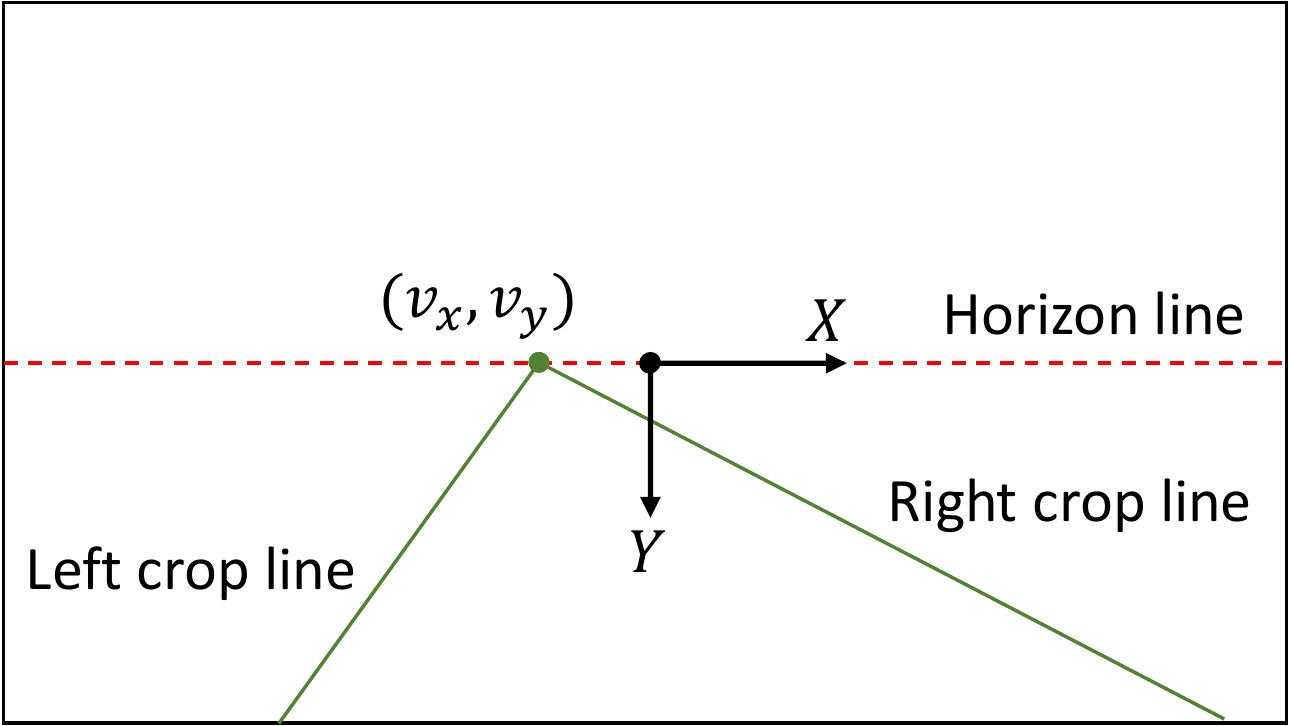}
\caption{Shows $(v_x, v_y)$ once the image and annotation lines have been rectified for roll and pitch.}
\figlabel{lr_intersection}
\end{figure}

\begin{equation}
    \eqlabel{heading_crop_lines}
    \begin{aligned}
        (A^l_X, A^l_Y, A^l_Z) + \lambda (\sin\phi, 0, \cos\phi)
    \end{aligned}
\end{equation}
for different values of $\lambda$. These points project to the image place at locations,

\begin{equation}
    \eqlabel{heading_projection}
    \begin{aligned}
        &\left(f\frac{A^l_X+\lambda\sin\phi}{A^l_Z + \lambda\cos\phi}, f\frac{0}{A^l_Z+\lambda\cos\phi}\right) \\
    \end{aligned}
\end{equation}

Vanishing point can be obtained by taking $\lim_{\lambda\to\infty}$,
\begin{equation}
    \eqlabel{heading_projection}
    \begin{aligned}
        &\lim_{\lambda\to\infty}\left(f\frac{A^l_X+\lambda\sin\phi}{A^l_Z + \lambda\cos\phi}, f\frac{0}{A^l_Z+\lambda\cos\phi}\right) \\
        &= (f\tan\phi, 0)
    \end{aligned}
\end{equation}

Thus, the heading $\phi$ can be obtained from the x-coordinate of the vanishing point, $v_x$ as, $\arctan(v_x/f)$. $v_x$ can be obtained from the intersection of the image of the left and the right crop row.

\subsection{Distance Ratio Estimation}
\seclabel{distance}
Assume that we have rotated the image to correct for the heading. Crop rows are now in the direction $(0, 0, 1)$. Let's assume that the left crop row goes through the point $(X_l, H, 0)$, and the right crop row passes through the point $(X_r, H, 0)$. Then the image of the left crop rows in the image plane is given by:

\begin{equation}
    \begin{aligned}
        \left( f\frac{X_l + \lambda_l\cdot0}{0 + \lambda_l\cdot 1}, f\frac{H + \lambda_l\cdot0}{0 + \lambda_l\cdot 1}\right) = \left( f\frac{X_l}{\lambda_l}, f\frac{H}{\lambda_l}\right) \\
    \end{aligned}
\end{equation}
The X-intercept of this line in the image plane is given by the value of $\lambda_l$, such that $f\frac{H}{\lambda_l}$ is equal to $h/2$ (where $h$ is the image height), \ie, $\lambda_l = f\frac{H}{h/2}$. Thus, the X-intercept of the left crop row, $l_x$ is $\frac{X_l\cdot h}{2\cdot H}$.

Similarly, the X-intercept for the right crop row, $r_x$ is $\frac{X_r\cdot h}{2\cdot H}$. Thus, the distance ratio, $d = \frac{X_l}{X_l+X_r}$ can be obtained via $\frac{l_x}{l_x+r_x}$.

\end{document}